# Intelligent Bearing Fault Diagnosis Method Combining Mixed Input and Hybrid CNN-MLP model

Sinitsin, V.*[1], Ibryaeva, O.[1], Sakovskaya V.[1], Eremeeva V.[1]

[1]School of Electronic Engineering and Computer Science, South Ural State University, Chelyabinsk, Russia

e-mail: sinitsinvv@susu.ru

**Abstract**

Rolling bearings are one of the most widely used bearings in industrial machines. Deterioration in the condition of rolling bearings can result in the total failure of rotating machinery. AI-based methods are widely applied in the diagnosis of rolling bearings. Hybrid NN-based methods have been shown to achieve the best diagnosis results. Typically, raw data is generated from accelerometers mounted on the machine housing. However, the diagnostic utility of each signal is highly dependent on the location of the corresponding accelerometer. This paper proposes a novel hybrid CNN-MLP model-based diagnostic method which combines mixed input to perform rolling bearing diagnostics. The method successfully detects and localizes bearing defects using acceleration data from a shaft-mounted wireless acceleration sensor. The experimental results show that the hybrid model is superior to the CNN and MLP models operating separately, and can deliver a high detection accuracy of 99,6% for the bearing faults compared to 98% for CNN and 81% for MLP models.

## 1 Introduction

As the rolling element bearings are key components of rotating machinery, their physical condition has a major impact on the safe and efficient operation of the equipment. Rolling bearing failures account for 30% or more of all failures in rotating machinery [1]. Accordingly, condition monitoring and intelligent diagnosis of bearings are considered critical aspects of system design and maintenance and have been the focus of widespread research effort over the past few decades [2]–[9].

In essence, fault diagnosis can be thought of as a pattern recognition problem related to the condition of rotating equipment. A common fault diagnosis method typically consists of two key steps: feature extraction (data processing) and fault classification. Vibration based signal processing is one of the most commonly used method for the first step, typically employing time-domain, frequency-domain, or time-frequency domain analysis.

Typically, a time-domain analysis calculates statistical parameters such as RMS, kurtosis, structural resonances [10] etc. Frequency-domain analysis is often advantageous, as it can readily isolate and identify key frequency components. A commonly used tool is the fast Fourier transform (FFT), as well as FFT based methods, spectrum analysis methods [11] etc. Time-frequency analysis is used to extend the capability of frequency-domain analysis to non-stationary vibration signals and includes such methods as the short-term Fourier transform [12], wavelet transform [13], empirical mode decomposition (EMD) of Hilbert–Huang transform (HHT) [14], EMD related methods etc. With the development of nonlinear dynamic theory, many entropy-based estimation methods provide useful alternative approaches for extracting the defect-related features hidden in vibration signals and applying them to fault detection for rolling bearings [15]–[17].

For the second step, the extracted features are used as inputs to machine learning techniques for the purposes of fault recognition. Numerous machine learning tools have been utilized. For example, the k-nearest neighbors (k-NN) method has been applied [18] to identify five different gear crack levels under different motor speeds and loads. Features extracted by HHT were also used by a k-NN classifier in [19]. Muralidharan et al. [20] adopted naive Bayes and Bayes network algorithms to implement fault diagnosis in combination with discrete wavelet transform for time–frequency features extraction. A support vector machine approach was utilized for fault diagnosis of rolling bearing in [21], [22] where features with most dominant fault information were extracted based on the concept of entropy. Hu et al. [23] use Random Forests to determine fault types, based on extracted multi-scale dimensionless indicators as fault features.

Artificial neural networks and especially deep neural networks have shown great potential in mechanical fault diagnosis [24]. Convolutional neural networks (CNN) have prevailed in bearing fault diagnosis in recent years. Zhang et al. [25] successfully trained a one-dimensional CNN with raw vibration signals. But, due to its key architectural features - local receptive fields, weight sharing, and sub-sampling in spatial domain - CNN is most suitable for processing 2-D data. In references [26], [27], and [28], time sequences of vibration signals were converted into time-frequency images by Wavelet analysis, HHT and FFT, respectively. Subsequently CNN was employed to exploit useful information from the images and recognize the fault patterns. A method of converting vibration signals in the time domain into a 2-D form, called a vibration image, is proposed in [3]. Again, CNN is used to identify bearing faults via vibration image classification.

It is becoming increasingly common to use data of different types within a single model. For example, Ma et al. [29] used features extracted in time domain and frequency domain in their PSO-SVM model for rolling bearing fault diagnosis. It is also becoming popular to create hybrid and collaborative techniques from existing machine learning algorithms. In [30], CNN is used to extract intrinsic fault features from the images (obtained from the continuous wavelet transform) which are then fed into a gcForest classifier. The analysis results demonstrated that the proposed hybrid deep learning model can achieve higher detection accuracy than CNN and gcForest operating in isolation.

Most of the above-mentioned studies are based on the data provided by the Case Western Reserve University (CWRU) Bearing Data Center [31] which has become a benchmark data set in the bearing diagnostics field. CWRU vibration data was collected using accelerometers, which were attached to the housing of a mechanism. It is the traditional approach for collecting diagnostic information from technological processes.

In [32] a wireless acceleration sensor is proposed which can be mounted directly on a rotating shaft. This sensor is able to measure angular and linear accelerations simultaneously, directly from the rotating shaft or gears. The construction of the sensor's moving part is a disk (a printed circuit board fixed on a hard base plate, in practice) rigidly mounted on the rotating shaft. The disk contains three one-axis accelerometers. Moreover, the accelerometers are mounted equidistant from the center with angle of 120 degrees between them. Furthermore, the sensitivity axes of the accelerometers are oriented tangentially to the rotating shaft (Fig. 1).

This method of mounting potentially provides increased sensitivity to defects [33], [34]. The data set used in this paper is unique, since it was obtained using a sensor mounted directly on the shaft.

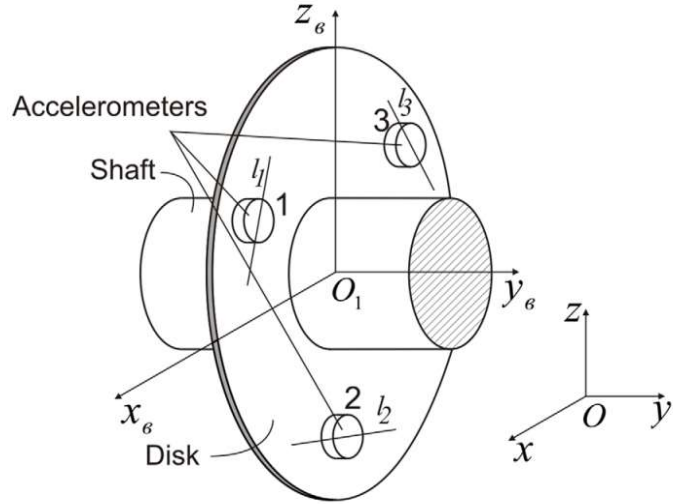

Fig. 1 Sensor model

Based on one of the state-of-the-art bearing fault diagnosis schemes (HHT-CNN), this paper proposes a novel intelligent bearing fault diagnosis method using mixed inputs. In the proposed hybrid model, the first stage (CNN) implements fault diagnosis from HHT images. The second stage (MLP - multi-layer perceptron) operates on the signal power at resonant frequencies. The proposed model therefore combines two NN architectures, using different data types as input – both numerical and images. As will be shown, the resulting model delivers higher diagnosis accuracy than either CNN and MLP in isolation.

The main contributions of this work can be summarized as follows:

1. A novel fault diagnosis method is proposed for simultaneously processing data of different types. This model combines MLP for numerical inputs and CNN for HHT images.

2. A new dataset is provided, obtained using a sensor mounted directly on the rotating shaft. This data set is described in Section 3 of the paper and available in the public domain [35]. We encourage other researchers to develop enhanced diagnostic methods applied to this unique data set.

3. Experiments on the dataset demonstrate that the proposed hybrid model is superior to CNN and MLP applied separately.

The remainder of this paper is structured as follows. Section 2 describes the proposed CNN-MLP hybrid model. Experimental validation is performed to evaluate the method in Section 3. Section 4 offers conclusions and future work.

## 2 Proposed intelligent fault diagnosis method

This section describes the hybrid fault diagnosis model that simultaneously processes data of different types. Fig. 1 depicts an overview of the proposed method.

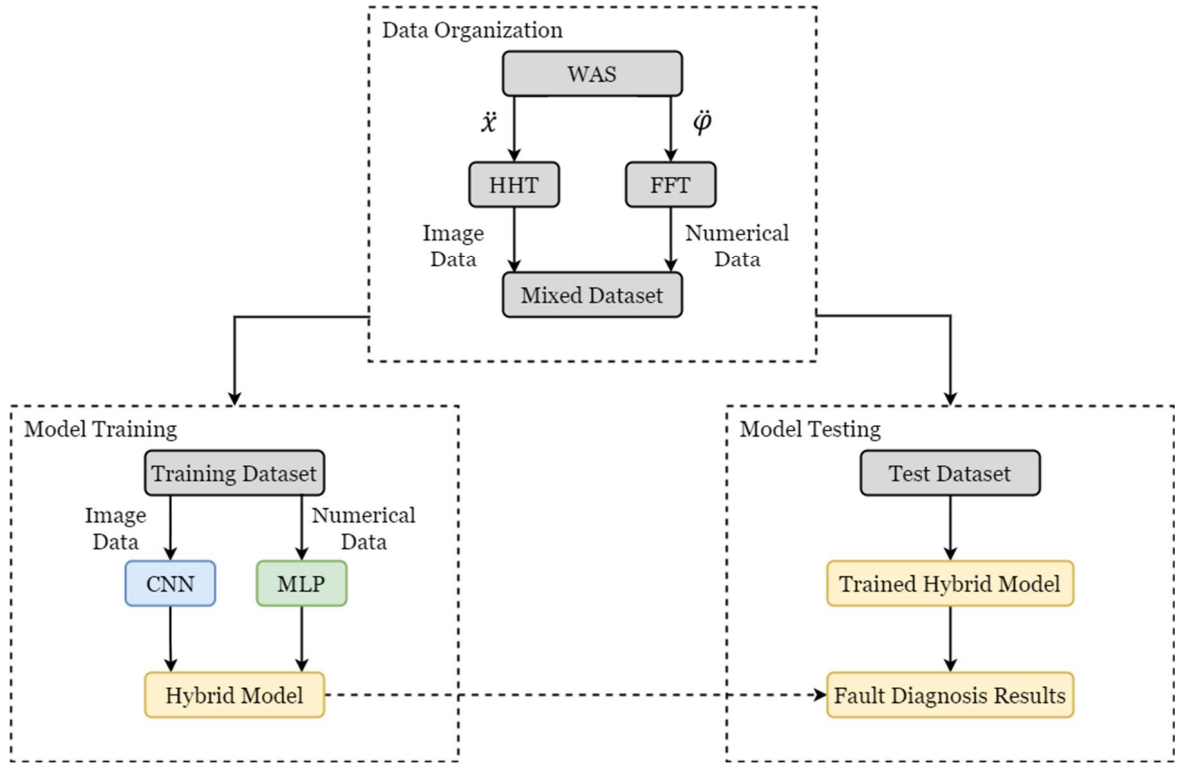

Fig. 1 Overview of the proposed method

The implementation of the method is as follows.

Step 1: WAS signals are converted to mixed input data. The linear acceleration signals (linear signals) are transformed into time-frequency images using HHT, which are suitable inputs for CNN. Similarly, angular acceleration signals (angular signals) are converted into numerical values for the signal power around the first (N1) and second (N2) frequencies of the shaft torsional modes. The values N1 and N2 serve as input data for the MLP.

Step 2: Model training using the data sets specific to each model type, resulting in a well-trained hybrid model.

Step 3: Model testing. Use the trained hybrid model to identify bearing faults based on the mixed input data.

## 2.1 Converting WAS signals to mixed input data

Signals of angular and linear accelerations (angular and linear signals, respectively) are obtained by means of the WAS [32], with a sampling rate of 31.175 kHz. Signals examples are presented in Fig. 2, where the shaft rotation frequency is 20Hz.

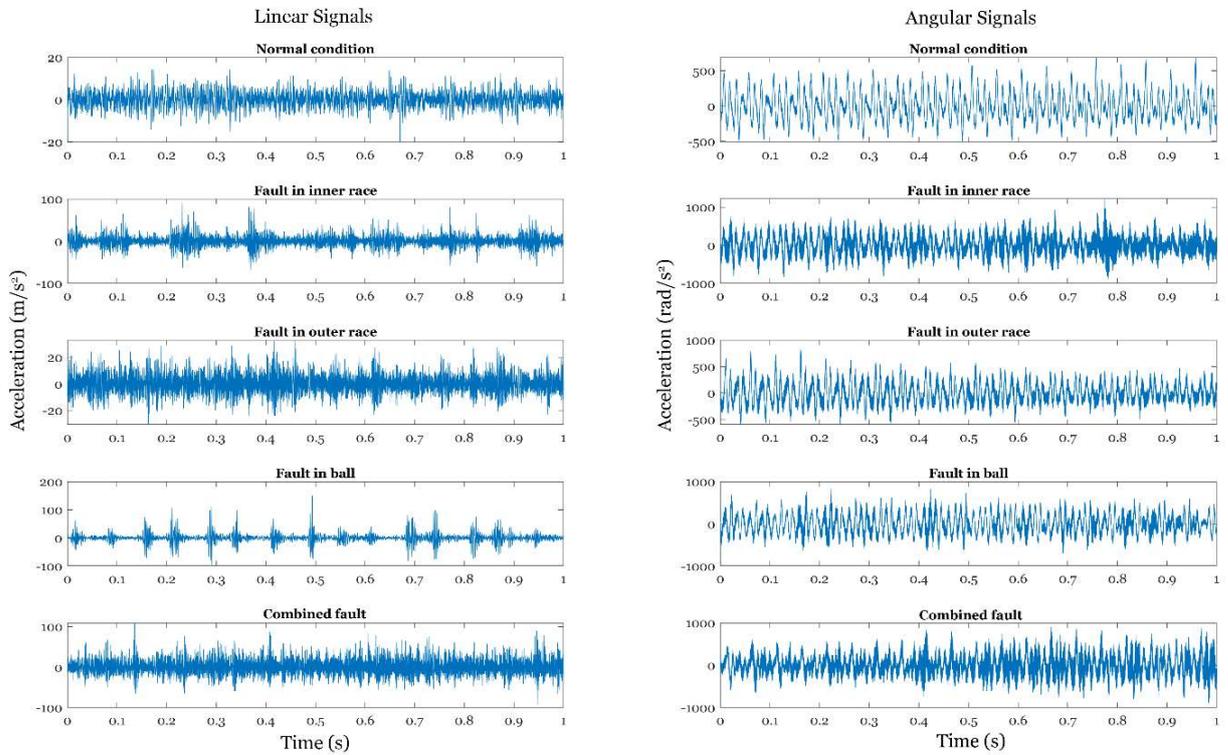

Fig. 2. Linear and angular signals

The signals-to-mixed input data conversion process is illustrated in Fig. 3.

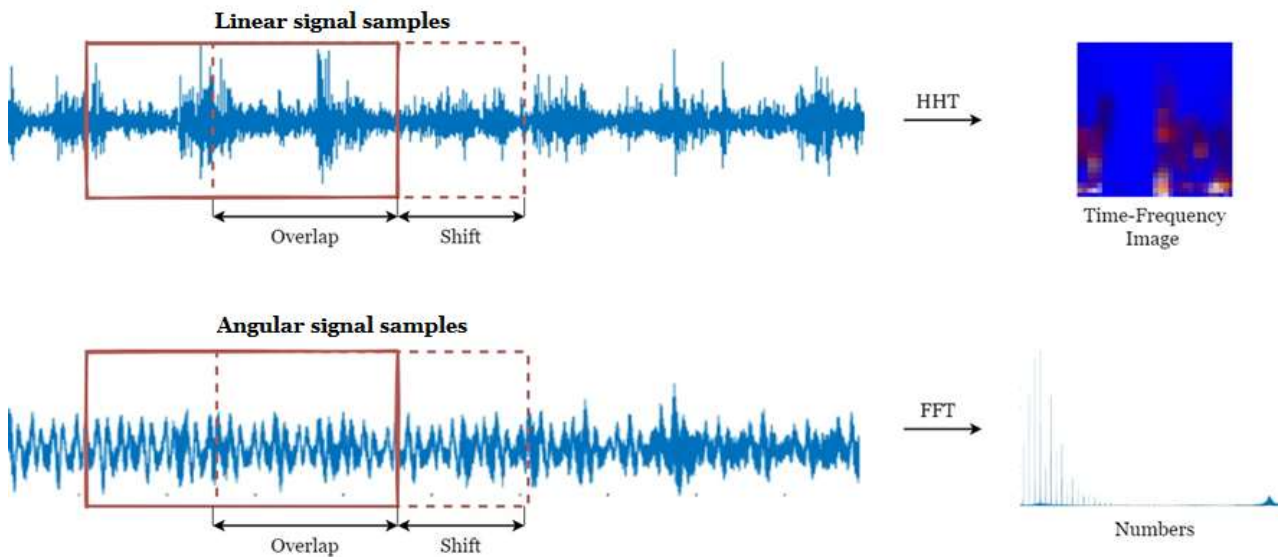

Fig.3 Signals-to-mixed input data conversion

Data is processed as a series of overlapping windows, where each window has duration of 125 ms (corresponding to 3,897 consecutive samples or 2.5 rotation periods), and where the offset between widows is 50 ms (corresponding to 1559 samples or 1 rotation period). HHT is used to transform each linear acceleration signal data window into a 32 x 32 time–frequency image to be processed by the CNN model. Simultaneously, FFT is applied to the corresponding angular acceleration data window to obtain N1 and N2. This process is repeated for subsequent overlapping data windows to produce the required data for training and testing.

Next, we describe the data processing in more detail.

### 2.1.1 Linear signal-to-image conversion

The Hilbert-Huang transform, which is a combination of the Empirical Mode Decomposition (EMD) and Hilbert transform, is a time–frequency analysis method designed for non-linear and non-stationary data [36]. It is often used in the analysis of bearing signals with complex frequency components [27].

Using EMD, an arbitrary signal can be adaptively decomposed into a collection of intrinsic mode functions (IMFs), and can be expressed as a sum of IMFs plus a residual term:

$$x(t) = \sum_{k} IMF_k(t) + r(t)$$

Following EMD, the Hilbert transform can be applied to each IMF separately. The Hilbert transform $y(t)$ of any signal $x(t)$ is defined as:

$$y(t) = \frac{1}{\pi} \int_{-\infty}^{\infty} \frac{x(t)}{t-\tau} d\tau$$

Utilizing $y(t)$ and $x(t)$ the associated analytical signal $z(t)$ is defined as:

$$z(t) = x(t) + iy(t) = a(t)e^{i\theta(t)}$$

where $a(t)$ is the envelope of the signal and $\theta(t)$ is the instantaneous phase. The instantaneous frequency can be calculated as the derivative of the instantaneous phase:

$$\omega(t) = \frac{d}{dt}\theta(t)$$

By performing the Hilbert transform on each IMF the original signal can be expressed as the real part ($\Re$) in the following form

$$x(t) = \Re\left(\sum_{j} a_j(t)e^{i\theta_j(t)}\right) = \Re\left(\sum_{j} a_j(t)e^{i\int \omega_j(t)dt}\right).$$

The above equation gives both the amplitude and the frequency of each component as a function of time. This time–frequency distribution of the amplitude is called the Hilbert spectrum.

Fig. 4 depicts the decomposition process of a linear signal from a bearing with and without a fault. Fig. 5 depicts the corresponding HHT spectra for the first three IMFs.

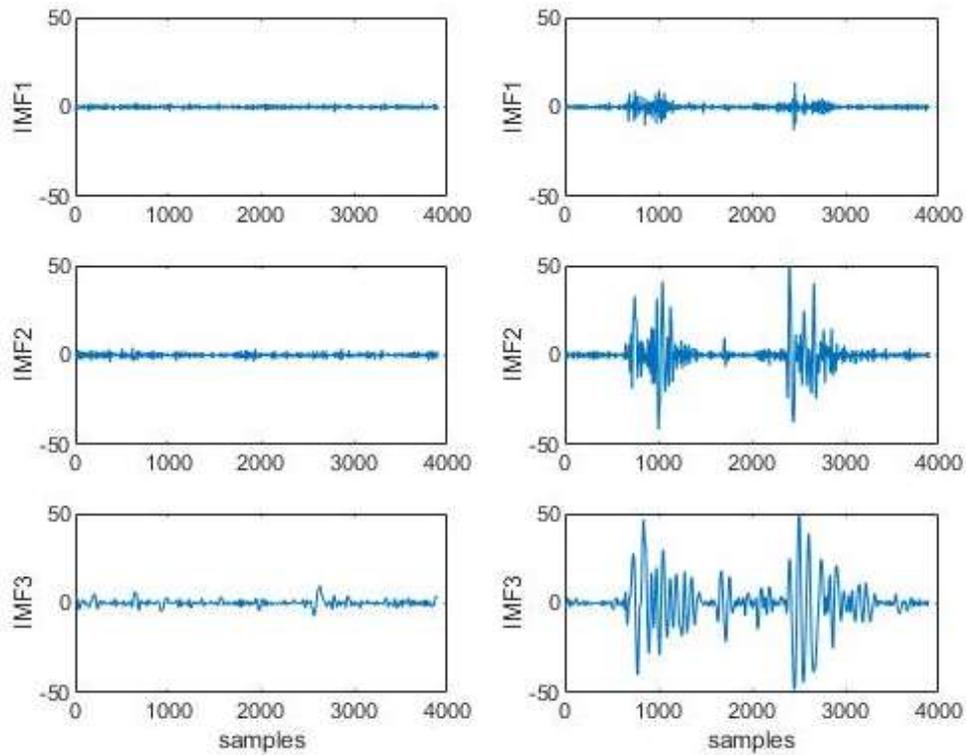

Fig. 4. IMFs for a healthy vibration signal (left) the bearing has a ball fault (right)

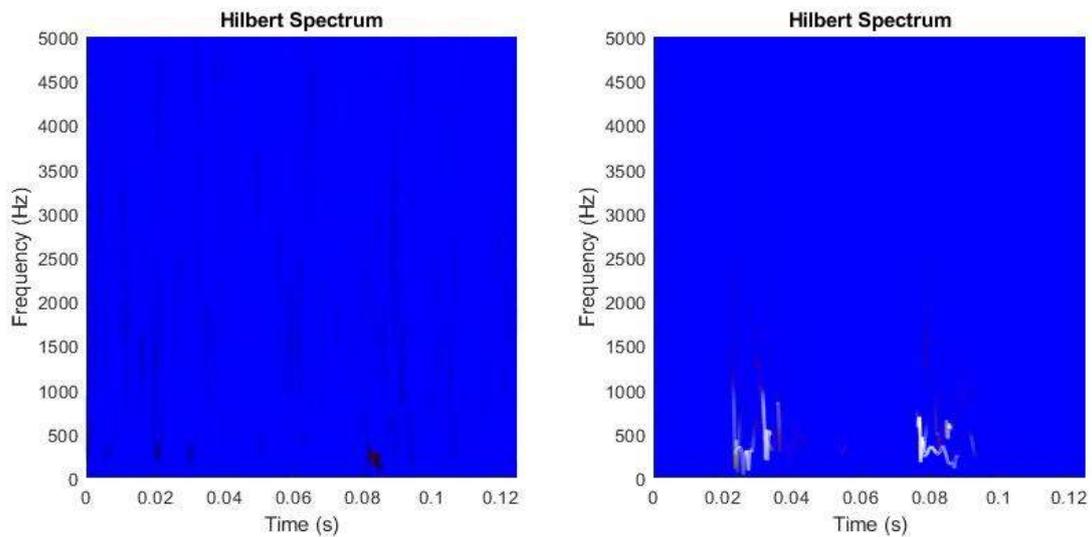

Fig 5. The Hilbert spectrum of the healthy vibration signal (left) and signal from a bearing with a ball fault (right)

In the traditional HHT method, the Hilbert spectrum of the whole signal is obtained by converting all obtained EMD decomposition results. However, good results with reduced computational cost have been obtained using only the first few IMFs to produce the Hilbert spectrum [37], [38]. Our numerous experiments have shown that it is most effective to use only the first three IMFs.

**2.1.2 Angular signals-to-numbers conversion**

The values N1 and N2 are based on an impulse model of defect behavior. Each defect in a bearing can generate a shock, which excites natural mechanical frequencies. Periodic shocks generate modulated frequency components around the natural frequencies including torsional modes (in the angular acceleration spectrum). Over time, the increasing bearing defects result in the strengthening of existing frequency components and the appearance of new ones. Accordingly, the analysis of angular acceleration signal power around the natural frequencies is useful for bearing fault detection.

Here, the torsional natural frequencies are found using a shock response spectrum (SRS) analysis. The SRS analysis contains three steps: (1) A hammer is used to generate a shock to the shaft. The shock response is measured by the WAS sensor which is fixed to the shaft. (2) The raw signals are decomposed into angular and linear accelerations based on a mathematical model of the sensor response. (3) The frequency peaks are found at the angular acceleration spectrum by FFT analysis. The resulting frequency peaks are the centers of signal power estimate bands. The SRS for the 3/4-inch shaft used in this work is shown in Fig 6, with data for each axis (angular and X-Y linear).

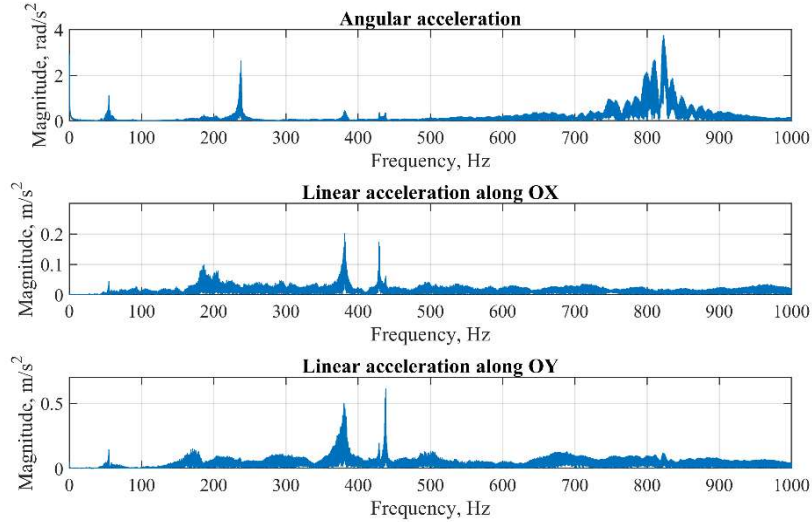

Fig 6. The shock response spectrum of the 3/4-inch shaft without rotating.

The SRS of the angular acceleration has two clear peaks at approximately 240 Hz and 820 Hz, corresponding to the first and second torsional natural frequencies of the 3/4-inch shaft. The values N1 and N2 are computed as the sum of the FFT components around the first and the second frequency peaks (Fig. 7). Values of the numbers N1 and N2 are associated with bearing defects.

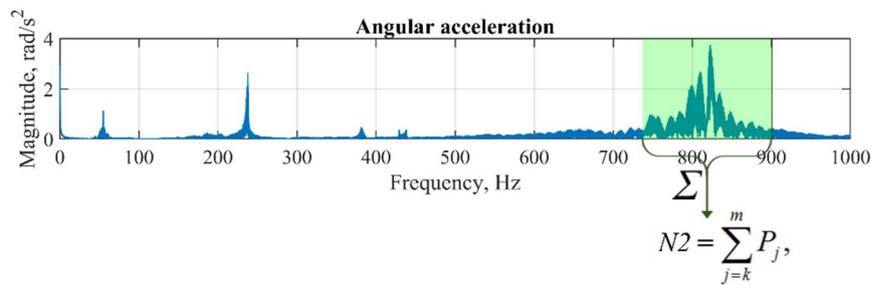

Fig. 7. Calculating the number N2

## 2.2 Hybrid CNN-MLP model

An overview of the developed hybrid model with mixed input is shown in Fig.8. This architecture is represented as a combination of MLP and CNN. Specifically, MLP is adopted to deal with numerical inputs, CNN is applicable to extract high-hierarchy features from structural data. MLP consists of multiple fully connected (FC) layers. CNN is composed of several convolution network units. After feature learning for different formats of data separately, the outputs of MLP and CNN are concatenated feature-wise to achieve the final classification results.

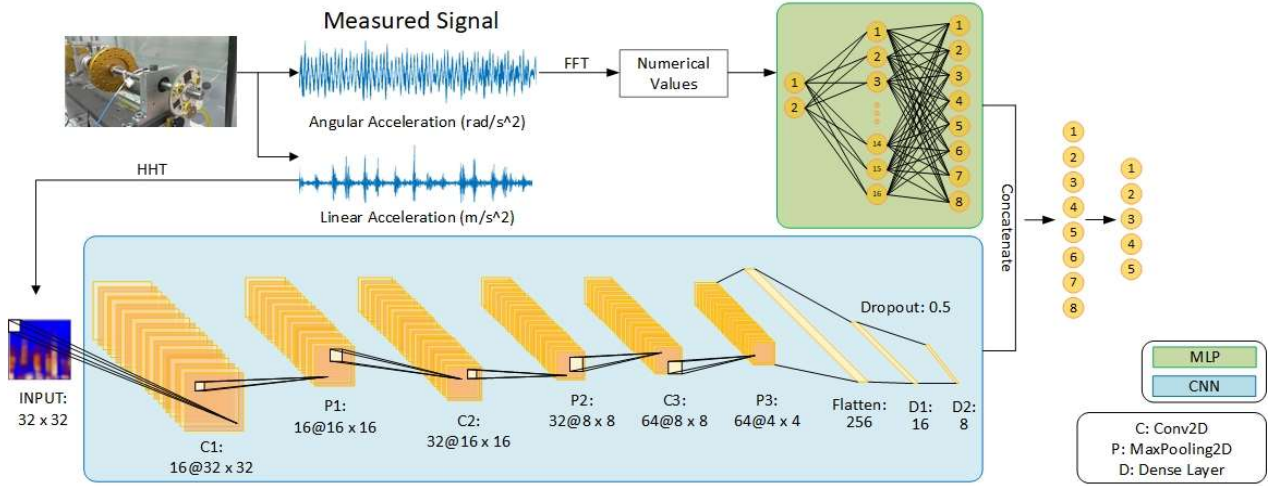

Fig.8. Hybrid CNN-MLP model with mixed input

The mixed input data for our hybrid model are image-number pairs found by linear and angular acceleration, respectively, in the same time interval. Before loading into the hybrid model, all input data (numbers N1, N2 and intensities of image pixels) are normalized to the range of [0,1].

## 3 Case study

In order to evaluate the performance of the proposed model for bearing fault diagnosis, we apply it to the data obtained on an experimental rig using the WAS-prototype.

### 3.1 Experimental rig

In the study, the experimental rig contains a 1-inch shaft supported by two bearing assemblies. The left bearing is defect-free while the on the right is the test bearing. The shaft is rotated by an AC motor which is fixed to the shaft by a jaw coupling adjacent to the defect-free bearing. A revolution indicator measures rotating shaft frequency. The WAS prototype is fixed at the end of the shaft near the test-bearing (Fig. 9). The WAS prototype contains three one-axis MEMS-accelerometers ADXL-001 (Analog Devices). A detailed description of the prototype, location and orientation of the accelerometers are described in [32].

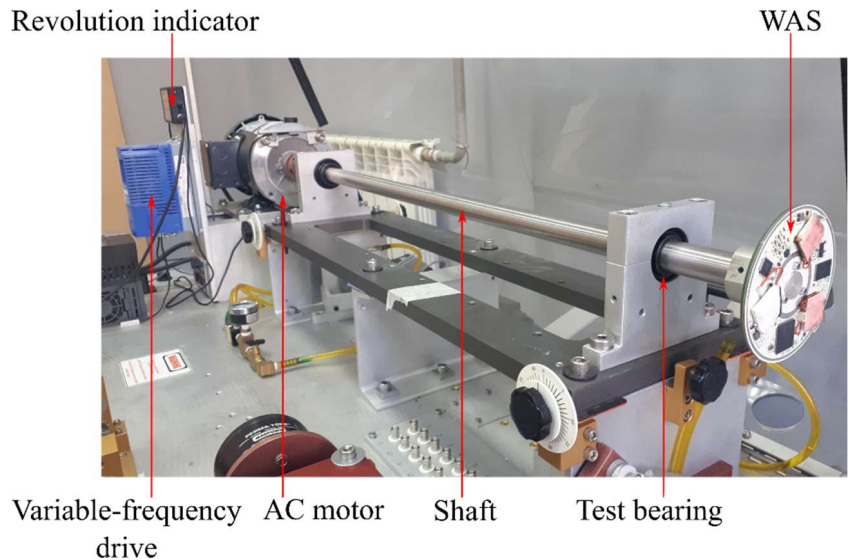

Fig.9. Experimental rig

A commercial bearing, the ER-16K from MB Manufacturing was used for both the test and defect-free devices. A rotary tool was used to apply mechanical damage to test-bearings, with three different fault types: damage to the inner race, outer race, and ball. A fourth bearing had all three faults applied. Thus, testing took place on a total of five test-bearings: one defect-free, three with single faults, and one with the three faults combined.

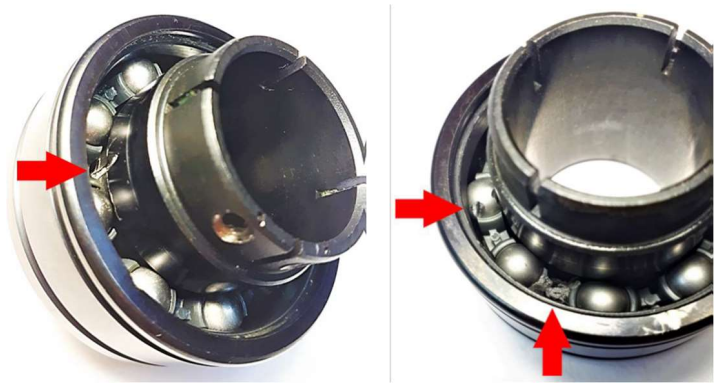

Fig.10. Test-bearings with inner race fault (left) and ball fault (right)

### 3.2 Data description and preliminary analysis

The linear and angular signals of all five bearings were collected with a sampling frequency of 31.175 kHz with the system operating at 1200 rpm (i.e. at 20 Hz).

The overlapping truncation method described in Section 2.1 was used to generate 27,900 HHT images from the linear signals. As shown in Fig. 3, a truncating window slides along the raw linear signal with a shift interval of 1,559 samples (which corresponds to one revolution of the shaft). The window size was 3,897 data points in this study (i.e. covering 2.5 shaft revolutions). HHT is used to transfer the selected linear signal data set into a 32 x 32 time–frequency image while FFT is applied to the corresponding angular signal data set to calculate N1 and N2.

Using the train_test_split() function from the data science library scikit-learn [39], we split the dataset into train and test subsets. The distribution of data sets between each class is shown in Table 1.

**Table 1.** Distribution of data sets for model development

| Fault type | Training data set | Validation data set | Test data set |
|---|---|---|---|
| Normal | 4036 | 705 | 839 |
| Inner race | 4056 | 696 | 828 |
| Outer race | 4018 | 746 | 816 |
| Ball | 3985 | 725 | 870 |
| Combined fault | 4062 | 686 | 832 |
| **Total** | 20157 | 3558 | 4185 |

In order to explore the data and evaluate the possibility of solving the classification problem, all 27,900 HHT images were visualized using the t-distributed stochastic neighbor embedding (t-SNE) [40] which preserves the original data structure.

Since our dataset is high dimensional (each image is characterized by 32*32*3=3072values) and t-SNE is computationally expensive, we perform principal component analysis (PCA) [41] before t-SNE to reduce the dimensionality of the data.

An essential aspect of applying PCA is the ability to estimate how many components are needed to adequately describe the data. This can be determined by looking at the cumulative explained variance ratio as a function of the number of components (Fig. 11). This curve quantifies how much of the total, 3072-dimensional variance is contained within the first N components. For example, we see that we need around 500 components to describe approximately 0.99 of the variance.

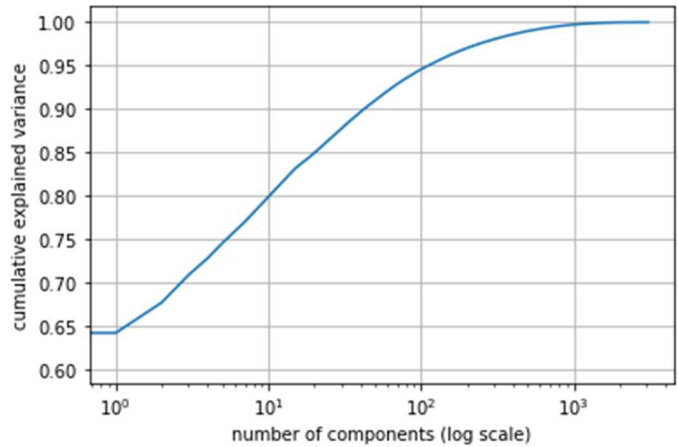

Fig. 11 Cumulative explained variance ratio as a function of the number of components

After reducing the number of features by PCA (to 506 components), we employ t-SNE to portray the resulting two-dimensional map. Fig. 12 shows the t-SNE visualization results of the HHT images. Each point in the Fig. 12 corresponds to one HHT image, so in total we have 27,900 points. Now each point, instead of 506 features, is characterized by only 2 features. The main advantage of t-SNE is the ability to preserve local structure. This means, roughly, that points which are close to one another in the high-dimensional data set will tend to be close to one another in the chart.

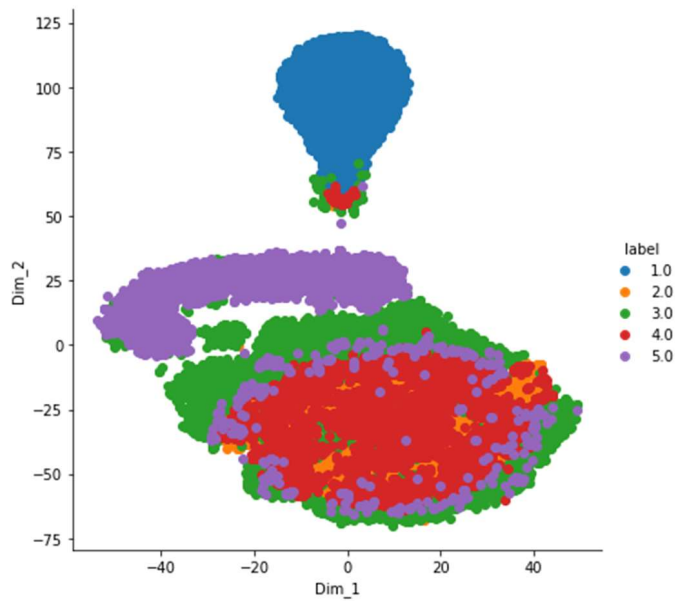

Fig. 12 Visualization of HHT images with t-SNE

1 – normal, 2 – inner race, 3 – outer race, 4 – ball, 5 – combined fault t-SNE visualization can be used to check for the presence of clusters in the data and to see if there is some order or some pattern in the dataset. It can aid our intuition about what we think we know about the domain we are working in. However, t-SNE is the unsupervised learning algorithm and it does not use any class label data in its work. In the above visualization, different colors result from metadata (label) embedding. If we turn off the color in Fig. 12, then we can clearly distinguish only three clusters, corresponding to a normal signal, a signal with a combined error, and signals with other types of defects. Thus, from any model that solves the problem of classifying our data set, we can expect good results in determining a normal signal, and a signal with a combined defect. We also expect more model errors in separating signals of the other three types (with inner race, outer race, ball faults), especially in recognizing signals with inner race and ball defects, which correspond to orange and red colors and which almost overlap each other in the Fig. 12.

### 3.3 Experimental results

**Table 2.** Detailed structure of the designed Hybrid CNN-MLP model

| CNN branch | | | | |
|---|---|---|---|---|
| **Input** | | 32 × 32 | | |
| **C1** | 16@ 3 × 3 | 16@32 × 32 | | |
| **P1** | 2 × 2 | 16@16 × 16 | | |
| **C2** | 32@ 3 × 3 | 32@16 × 16 | | |
| **P2** | 2 × 2 | 32@ 8 × 8 | | |
| **C3** | 64@ 3 × 3 | 64@ 8 × 8 | | |
| **P3** | 2 × 2 | 64@ 4 × 4 | MLP branch | |
| **Flatten** | 256 neurons | | **Input** | numbers N1, N2 |
| **D1** | 16 neurons + dropout | | **D1** | 16 neurons |
| **D2** | 8 neurons | | **D2** | 8 neurons |
| Concatenation | | | | |
| **Dense layer** | | | 8 neurons | |
| **Output** | | | 5 neurons | |

A trail-and-error analysis was carried out to determine the Hybrid CNN-MLP model parameters, as showed in Table 2. Layer C1 contained 16 convolution kernels with a size of 3×3 and outputted 16 feature maps with a size of 32×32 (we used zero padding to have the same size of output feature-maps). Layer P1 (2 ×2) outputted 16 pooling maps with a size of 16×16. Layer C2 produced 32 feature maps with a size of 16×16 and layer P2 provided 32 pooling maps with a size of 8×8. After a similar construction of layers C3, P3, we have 64 pooling maps with a size of 4×4 which than flatten in a layer with 256 neurons. An output from the flatten level is passed to two Dense layers D1, D2 with the dropout rate 0,5 between them.

On the other hand, N1 and N2 are the inputs to the MLP branch with two dense layers of 16 and 8 neurons, respectively. The concatenated outputs of the two models are sent to the fully connected layer of 8 neurons and then to an output layer of 5 neurons, which performs data classification (i.e. which generate 0/1 outputs corresponding to the five possible diagnostic outputs: no fault, each of the three individual faults, and the combined fault.

Adam optimizer [42] with an initial learning rate of 1e-4 was adopted to optimize the model parameters. All activation functions were the Relu function, the batch size was 20, and the loss function was Categorical Cross entropy function.

Initially, the number of epochs was set to 200 but we used Keras callbacks: EarlyStopping and ModelCheckpoint and due to this, the number of training epochs was less. These callbacks allow us to monitor the value of the loss function on the validation set and, if it starts to increase, stop the process of training the network. We used the value of the "patience" parameter equal to 50 epochs, i.e. we saved the best neural network (with the smallest value of the loss function on the validation set) and if there was no improvement in this value for 50 epochs, we stopped training and took the previously saved best model.

Learning curves and confusion matrices are shown in Fig. 13. Learning curves show how the values of accuracy and loss on the training and validation datasets changed during training process. Here the training of the Hybrid CNN-MLP model was stopped at the 112th training epoch, because there was no decrease in the value of the loss function on the validation set during 50 epochs. The model obtained at the 62nd training epoch was taken as the model with the lowest value of the loss function on the validation set. Further, a test dataset, which did not participate in the training process in any way, was given to this model. The results of processing the test dataset are well represented by the confusion matrix on the right in the Fig. 13.

The confusion matrix is a two-dimensional array comparing predicted category labels to the true label. Entry $i, j$ in the confusion matrix is the number of observations actually in group $i$, but predicted to be in group $j$.

As we can see from the confusion matrix for the Hybrid model, all normal signals, except one, were correctly recognized by the network as normal. One was mistakenly classified as a ball defect signal. Similarly, all signals with a defect in the inner ring, with the exception of two (erroneously assigned to classes Outer and Ball), were correctly classified. The signals with a ball defect presented the greatest difficulty for the model, 7 of them were mistakenly assigned to the class Inner. In total, the Hybrid model made 15 errors (the sum of the off-diagonal elements). Recall that the size of the test dataset was 4185 samples. Thus, the accuracy of the model was (4185-15) / 4185 = 0.9964 = 99.64%.

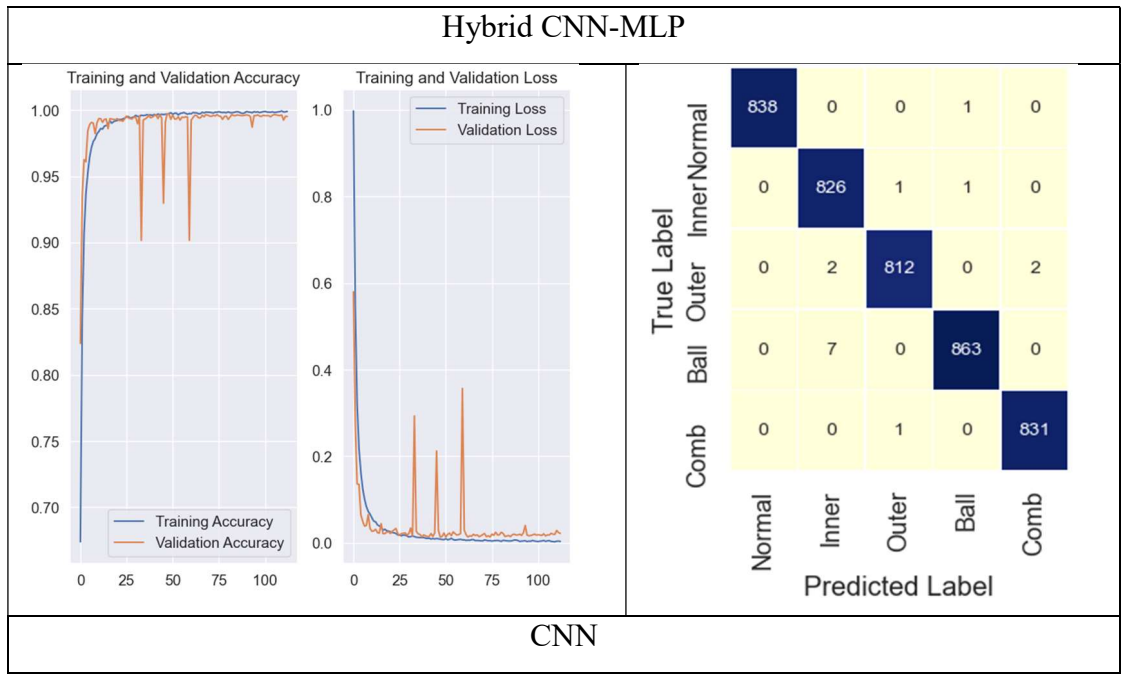

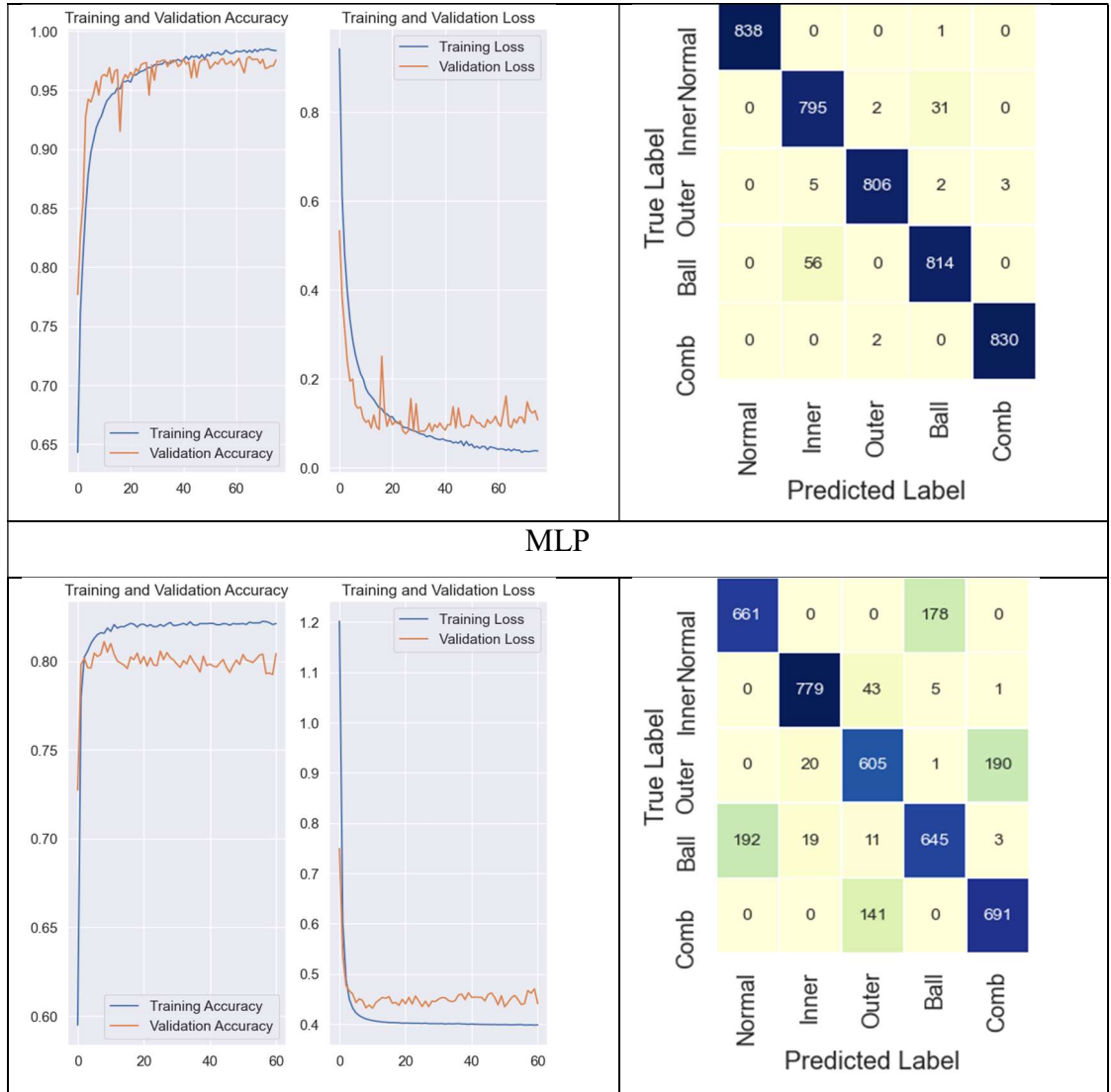

Fig. 13 Learning curves and confusion matrices of the fault detection results

Similarly, we can analyze the learning curves and confusion matrices for CNN and MLP models. We also stopped training if during 50 epochs there was no decrease in the value of the loss function on the validation dataset, and then we evaluated the accuracy of the model on the test dataset. For the CNN model, the greatest difficulty was also presented by signals with a ball defect, 56 of which were erroneously attributed to the Inner class. In turn, 31 signals of Inner class were erroneously assigned to the Ball class. We anticipated difficulties in the separation of Inner and Ball classes earlier in Section 3.2 from the analysis of Fig. 12 obtained using t-SNE for HHT images.

The MLP model works only with N1 and N2, and its behavior is completely different. It confuses normal with ball defect, as well as outer ring with combined defect. Accordingly, the MLP and CNN models successfully complement each other and the Hybrid model generates significantly fewer errors.

The standard metric of accuracy simply shows the proportion of correct responses from the model. If we are interested in a more detailed analysis of its work, then it is appropriate to examine the precision and recall metrics.

Precision is defined as the fraction of relevant instances among all retrieved instances. Recall, sometimes referred to as sensitivity, is the fraction of retrieved instances among all relevant instances. A perfect classifier has precision and recall both equal to 1. Precision and recall are combined together into the f1-score:

$$f1 = 2 * \frac{Precision * Recall}{Precision + Recall}$$

Summary of the precision, recall, f1-score, accuracy for each class and each model is given in the classification report of Table 3.

As can be seen, the Inner race class is the most difficult for the CNN model, scoring a precision of 0.93. This is consistent with the confusion matrix results where 56 examples of the Ball class were erroneously assigned to the Inner race class.

By contrast, the MLP model performs well in identifying signals of the Inner race class. We can see that the values of all metrics for this class are in the range of 94-95%, significantly exceeding the corresponding values for the other fault classes. These strong results are perhaps explained by the nature of the fault and the WAS sensor location. The inner race is rigidly bound to the shaft, and hence is close to the WAS in terms of the kinematic chain.

**Table 3.** Classification report for the models

|  | **Hybrid CNN-MLP** | | | **CNN** | | | **MLP** | | |
|---|---|---|---|---|---|---|---|---|---|
|  | precision | recall | f1-score | precision | recall | f1-score | precision | recall | f1-score |
| Normal | 1.00 | 1.00 | 1.00 | 1.00 | 1.00 | 1.00 | 0.77 | 0.79 | 0.78 |
| Inner race | 0.99 | 1.00 | 0.99 | 0.93 | 0.96 | 0.94 | 0.95 | 0.94 | 0.95 |
| Outer race | 1.00 | 1.00 | 1.00 | 1.00 | 0.99 | 0.99 | 0.76 | 0.74 | 0.75 |
| Ball | 1.00 | 0.99 | 0.99 | 0.96 | 0.94 | 0.95 | 0.78 | 0.74 | 0.76 |
| Combined fault | 1.00 | 1.00 | 1.00 | 1.00 | 1.00 | 1.00 | 0.78 | 0.83 | 0.80 |
| **Accuracy** |  | **1.00** |  |  | **0.98** |  |  | **0.81** |  |

### 4 Conclusions and future work

This paper proposes a new method for bearing fault diagnosis based on a Hybrid CNN-MLP model. The model simultaneously processes input data of different types and consists of two blocks: MLP to process numerical inputs and CNN to process HHT images. The unique dataset obtained using a sensor mounted directly on the shaft is presented. This data acquisition method is more sensitive to bearing defects. We provide open access to this dataset and encourage other scientists to use it.

It is shown that the proposed hybrid model is superior to the CNN and MLP models in isolation and can produce a high detection accuracy of 99,6% for bearing faults compared to 98% for CNN and 81% for MLP models. All experiments in this work were carried out for one shaft rotation frequency. In future studies, it is planned to test how the proposed technique works in the case of signals with different shaft speeds.

**Acknowledgments**

The research was funded by RFBR and Chelyabinsk Region, project number 20-48-740031.

**Availability of data**

The data that support the findings of this study are available at [35].